\documentclass[lettersize,journal]{IEEEtran}
\usepackage{amsmath,amsfonts}
\usepackage[noend]{algorithmic}
\usepackage{algorithm}
\usepackage{array}
\usepackage[caption=false,font=normalsize,labelfont=sf,textfont=sf]{subfig}
\usepackage{textcomp}
\usepackage{stfloats}
\usepackage{url}
\usepackage{verbatim}
\usepackage{graphicx}
\usepackage{cite}
\usepackage{booktabs} 
\usepackage{multirow}
\usepackage{xcolor}
\usepackage[super]{nth}
\usepackage{tikz}

\usepackage{soul}
\usepackage{enumitem}

\usepackage[switch]{lineno}

\begin{document}


\title{PSViT: A Methodology for Structurally Pruning Spiking Vision Transformers}

\author{Rachmad Vidya Wicaksana Putra,~\IEEEmembership{Member,~IEEE,} Achyuta Muthuvelan, Alberto Marchisio,~\IEEEmembership{Member,~IEEE,} and Muhammad Shafique,~\IEEEmembership{Senior Member,~IEEE} 
\thanks{Rachmad Vidya Wicaksana Putra, Achyuta Muthuvelan, and Alberto Marchisio are with eBRAIN Lab, Division of Engineering, New York University (NYU) Abu Dhabi, United Arab Emirates (UAE); (e-mail: rachmad.putra@nyu.edu, am12729@nyu.edu, and alberto.marchisio@nyu.edu). \\
Muhammad Shafique is the Director of eBRAIN Lab, Division of Engineering, New York University (NYU) Abu Dhabi, United Arab Emirates (UAE); 
(e-mail: muhammad.shafique@nyu.edu).}
}


\maketitle


\begin{abstract}
Spiking Vision Transformer (SViT) models are promising low-power ViT models for solving vision-based tasks with state-of-the-art performance.
However, their large sizes limit their deployments for resource-constrained embedded platforms, underscoring the needs of model compression.
One of prominent compression techniques is pruning, and the state-of-the-art works employ unstructured pruning techniques to compress SViT models.
Such techniques require specialized hardware architectures tailored for the sparsity patterns to maximize their efficiency benefits, making this approach not scalable.
To address this, we propose \textit{\textbf{PSViT}}, a novel methodology to perform structured pruning on SViT models, hence making it possible to efficiently accelerate their inference using the existing and widely-used computing architectures.
To do this, PSViT employs several key steps: \textit{uniform channel-wise filter pruning} to structurally eliminate the non-significant weights, \textit{sensitivity analysis} to evaluate the impact of channel-wise pruning of individual layer on accuracy and network size, as well as \textit{fine-grained channel-wise pruning} based on the sensitivity analysis and the given network architecture.     
Experimental results show that PSViT effectively obtains 22.4\% memory saving through single-shot pruning, while maintaining high accuracy within 3\% (70.3\% without fine-tuning and 72.8\% with fine-tuning) from the original non-pruned SViT model (73.3\%) on the ImageNet-1K.
These results also show that the PSViT methodology advances the effort in enabling efficient SViT deployments on resource-constrained applications. 
\end{abstract}


\section{Introduction}

In recent years, Vision Transformers (ViTs) have demonstrated state-of-the-art performance across a wide range of visual recognition tasks~\cite{dosovitskiy2020image, touvron2021training, Ref_Khan_SurveyViT_CSUR22}, outperforming conventional artificial neural networks (ANNs) due to their capacity to model long-range dependencies via self-attention mechanisms.
These advancements make the ViT deployments for diverse application use-cases highly desirable.
However, efficient deployments of ViT models are very challenging due to their large sizes and complex computations.
Recently, Spiking Neural Networks (SNNs) have been leveraged for developing alternative low-power ViT models, called \textit{Spiking Vision Transformers (SViTs)}~\cite{Ref_Zhou_Spikformer_ICLR23, Ref_Yao_SpikeDrivenTransformer_NeurIPS23, Ref_Yao_SpikeDrivenTransformer2_ICLR24}.  
SViTs combine the architectural expressiveness of conventional ViTs with the sparse event-based computation of spiking models.
However, despite their advantages, SViTs remain memory-intensive due to their huge number of parameters, creating significant obstacles for embedded deployments.

To compress the SNN-based models, previous works have proposed different methods, such as neuron removal~\cite{Ref_Putra_FSpiNN_TCAD20}, quantization~\cite{Ref_Rathi_PruneQuantizeSNN_TCAD18, Ref_Sorbaro_OptimSNN_FNINS20, Ref_Zou_MedianQuant_ISCAS20, Ref_Putra_QSpiNN_IJCNN21, Ref_Putra_QSViT_IJCNN25}, and weight pruning~\cite{Ref_Rathi_PruneQuantizeSNN_TCAD18}.
Pruning is one of the prominent compression methods since it can substantially compress model size, while undermining accuracy.
Therefore, pruning should be performed carefully, so that the accuracy degradation is still acceptable. 
While pruning has been extensively studied in traditional ANNs and ViTs~\cite{han2015deep, li2017pruning, marchisio2018prunet, zheng2022savit, yu2022width}, most methods require retraining to recover performance, which adds compute overhead and assumes access to labeled training data.
Moreover, pruning methods in the ANN domain cannot be employed directly for SNN-based models (e.g., SViTs) due to significant differences in the neuronal and synaptic operations.
Therefore, \textit{\textbf{the targeted research problem} is how we can prune SViT in single-shot to effectively reduce their memory footprints, while maintaining high accuracy and avoiding the excessive retraining overhead of iterative pruning methods}.
A solution to this problem may enable efficient SViT deployments on resource-constrained embedded AI applications with reduced development time.

\subsection{State-of-the-Art Pruning Methods for SViTs and Their Limitations}

Currently, the advancements of SViT models are still focusing on achieving high accuracy~\cite{Ref_Zhou_Spikformer_ICLR23, Ref_Yao_SpikeDrivenTransformer_NeurIPS23, Ref_Yao_SpikeDrivenTransformer2_ICLR24}.
Therefore, only several state-of-the-art target at pruning SViT models.
Recent work proposes the spatial-temporal spiking feature pruning (STSFP) method~\cite{Ref_Zhou_STSFP_TCDS25}, and another work proposes sparsification with timestep-wise anchor token and dual alignments (STATA)~\cite{Ref_Zhuge_TokenSparse_ICML24}, hence indicating that these works aim for input data/token sparsification instead of weight sparsification.
Consequently, they do not consider compressing the model sizes, hence their memory requirements remain the same as the original ones. 
Meanwhile, another work called Sparsespikformer~\cite{Ref_Liu_Sparsespikformer_ICASSP24} leverages the lottery ticket hypothesis for pruning tokens and weights.
However, these works employ the unstructured pruning, which requires specific hardware to achieve efficient computation for exploiting the sparsity patterns and compensating data indexing overheads, hence limiting its applicability for model acceleration on widely-used computing platforms (e.g., commodity neuromorphic hardware).
Moreover, STSFP and Sparsespikformer have not considered large and complex datasets like ImageNet-1K~\cite{deng2009imagenet}, which is crucial for real-world environments and benchmarking against state-of-the-art performance from the DNN domain. 
Hence, \textit{an alternative pruning method is required to compress any given SViT models while preserving high accuracy}.

\subsection{Motivational Study}

\begin{figure}[t]
\centering
\includegraphics[width=0.9\linewidth]{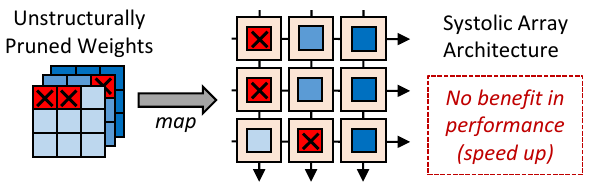}
\vspace{-0.3cm}
\caption{Limitations of mapping unstructurally-pruned weights on systolic array-based architecture, a widely-used hardware architecture for SNNs. No benefit in performance (speed-up) is obtained as the processing latency for completing all operations with unstructurally-pruned weights remains the same to that of the structurally-pruned ones.}
\label{Fig_Unstructured}
\vspace{-0.5cm}
\end{figure}

There are limitations of the unstructured pruning approach employed by state-of-the-art, as discussed in the following.
\begin{itemize}[leftmargin=*]
    \item \textit{Hardware mapping challenges:} 
    The pattern of zero entries in the pruned weights is typically random, hence making it challenging to efficiently map weights on the widely-used hardware platforms, such as commodity neuromorphic hardware~\cite{Ref_Demler_Akida_Linley19}\cite{Ref_Posey_Akida}. 
    \item \textit{Suboptimal performance efficiency benefits:} 
    Running unstructurally pruned networks on the widely-used computing platforms with regular structure, such as systolic array~\cite{Ref_Basu_SNNicSurvey_CICC22}\cite{Ref_Putra_SoftSNN_DAC22}, cannot maximize the performance efficiency benefits; see Fig.~\ref{Fig_Unstructured}.
    \item \textit{Specialized accelerators requirements:} 
    The unstructured pruning method typically has additional indexing policy to store positions of non-zero values, hence requiring specially designed hardware accelerators to maximize the performance efficiency benefits.
\end{itemize}
Based on these limitations, we identify the related research challenges to address below.
\begin{itemize}[leftmargin=*]
    \item SViT compression should employ the single-shot structured pruning method that can benefit from the widely-used hardware architectures (e.g., systolic array) without incurring high overhead of iterative pruning with multiple cycles of fine-tuning.
    \item The pruning method should facilitate different types of layers in SViT models, including convolutional (CONV), linear/fully-connected (FC), and attention layers. 
    \item The pruning method should employ an effective technique to evaluate and decide which weights to prune, considering the structure of filters and the hardware architecture. 
\end{itemize}

\subsection{Our Novel Contribution}

\begin{figure*}[t]
\centering
\includegraphics[width=\textwidth]{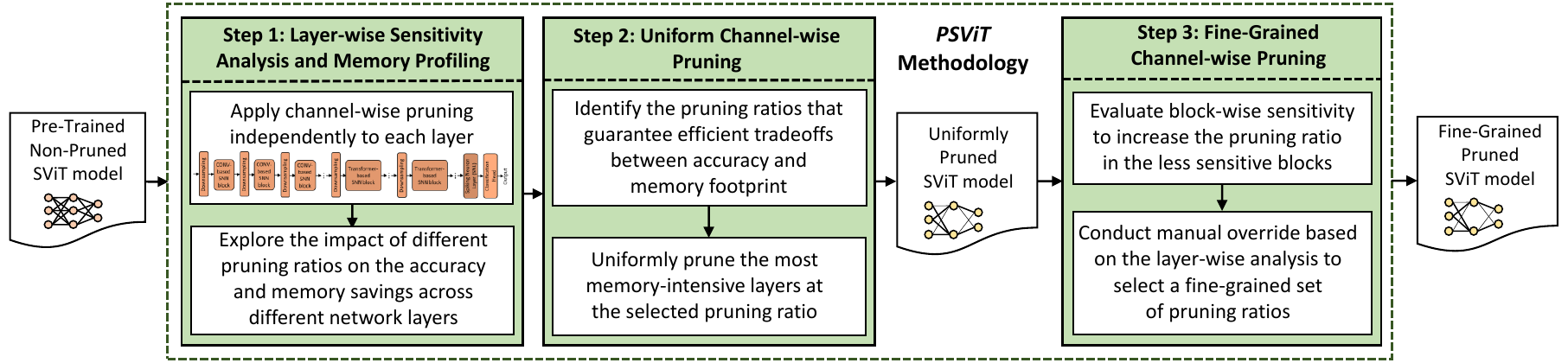}
\vspace{-0.6cm}
\caption{Overview of our proposed PSViT methodology, showing its main steps.}
\label{Fig_PSViT}
\vspace{-0.2cm}
\end{figure*}

To address the targeted problem and the research challenges, we propose \textit{\textbf{PSViT}, a novel methodology that effectively performs single-shot structured pruning for compressing SViT models, while maintaining high accuracy}.
It is also the first work that performs structured pruning with a single-shot approach for compressing SViTs. 
To achieve this, PSViT employs the following key steps (overview in Fig.~\ref{Fig_PSViT}).
\begin{itemize}[leftmargin=*]
    \item \textbf{Layer-wise Sensitivity Analysis and Memory Profiling (Section~\ref{Sec_PSViT_Sensitivity}):} 
    This step investigates the sensitivity of each network layer to channel-wise pruning across different pruning percentages, while evaluating the corresponding memory savings.
    \item \textbf{Uniform Channel-wise Pruning (Section~\ref{Sec_PSViT_UniformPrune}):} 
    This step identifies the pruning ratios that lead to notable memory savings while preserving high accuracy, then leverages this analysis to uniformly prune large layers.  
    \item \textbf{Fine-Grained Channel-wise Pruning (Section~\ref{Sec_PSViT_FinePrune}):} 
    This step first investigates the sensitivity of each network block, and then increases the pruning ratio in the less sensitive blocks.
    Afterward, it performs manual override based on the layer-wise sensitivity analysis for fine-grained pruning settings.
\end{itemize}

\textbf{Key Results:} 
We evaluate PSViT methodology through Python implementation based on SpikingJelly library~\cite{Ref_Fang_SpikingJelly_SciAdv23} and run it on the Nvidia RTX 4090 Ti multi-GPU machine. 
Experimental results show that PSViT effectively saves 22.4\% of memory footprint, while preserving high accuracy within 3\% from the original non-pruned SViT models considering the ImageNet-1K dataset~\cite{Ref_Deng_ImageNet_CVPR09}.

\section{Background and Related Work}

\subsection{Spiking Neural Networks (SNNs)}

Spiking Neural Networks (SNNs) are a class of neural models inspired by the discrete and event-driven signaling behavior of biological neurons~\cite{maass1997networks}. 
Unlike conventional artificial neural networks (ANNs), which operate with continuous-valued activations, SNNs transmit information via binary spikes over timesteps~\cite{Ref_Mozafari_SpykeTorch_FNINS19}, making them inherently temporally sparse. 
This event-driven computation leads to significant advantages in terms of energy efficiency, especially when deployed on neuromorphic hardware such as Intel's Loihi~\cite{Ref_Davies_Loihi_MM18} or IBM's TrueNorth~\cite{Ref_Akopyan_TrueNorth_TCAD15}.
The spatiotemporal dynamics of SNNs allow them to naturally encode and process time-varying signals, leveraging certain neural coding~\cite{Ref_Diehl_STDPmnist_FNCOM15, Ref_Park_BurstSNN_DAC19, Ref_Park_T2FSNN_DAC20, Ref_Putra_TopSPark_IROS23}. 
However, their non-differentiable spike dynamics pose challenges for gradient-based training, motivating the development of surrogate gradient methods~\cite{neftci2019surrogate}. 
Given those characteristics, SNNs are particularly suitable for energy-efficient applications where energy constraints are paramount (e.g., edge computing)~\cite{Ref_Putra_SpikeNAS_TAI25}\cite{Ref_Minhas_SurveyNC_Access25L}. 
Moreover, their compatibility with neuromorphic hardware makes them promising candidates for biologically-plausible and low-power AI intelligence.

\vspace{-0.2cm}
\subsection{Spiking Vision Transformers (SViTs)}

\begin{figure*}[t]
\centering
\includegraphics[width=\textwidth]{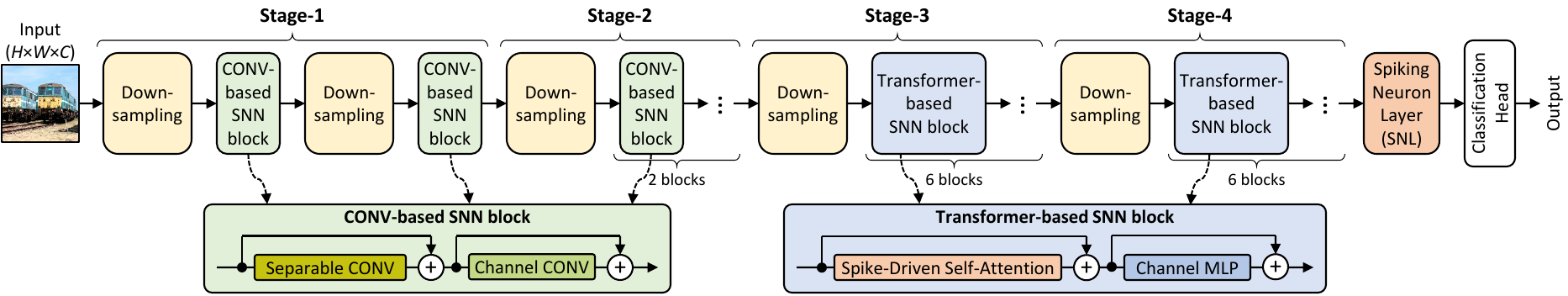}
\vspace{-0.6cm}
\caption{Architecture of the SDTv2 model, based on studies from~\cite{yao2024spikedriven}.}
\label{Fig_SDTv2}
\vspace{-0.3cm}
\end{figure*}

Recently, researchers equip ViTs with SNN components and operations to bridge the gap between the performance (accuracy) of ViTs and the energy efficiency of SNNs, known as Spiking Vision Transformers (SViTs). 
These architectures combine spiking neurons with transformer-based attention mechanisms to create models that are both temporally sparse and capable of complex reasoning. 
Recent SViT models such as Spikformer~\cite{Ref_Zhou_Spikformer_ICLR23}, Spike-Driven Transformer (SDT)~\cite{Ref_Yao_SpikeDrivenTransformer_NeurIPS23}, and Spike-Driven Transformer v2 (SDTv2)~\cite{Ref_Yao_SpikeDrivenTransformer2_ICLR24}.
These models usually have similar backbone architecture, including \textit{linear layers}, \textit{convolutional (CONV) layers}, and \textit{transformer-based attention layers}.
Sometimes, they also employ a downsampling block to reduce the computational burden.
These models are usually deep, having more than 5 layers to effectively process complex task/dataset like the ImageNet-1K.
Hence, multiple layers are usually grouped into a \textit{block}. 
Meanwhile, multiple blocks are also grouped into a \textit{stage}. 
Therefore, applying a pruning method to one model typically makes it applicable to other models.
In this work, we consider the state-of-the-art SDTv2. 
It has 4 stages, where there are 2 sets of downsampling and CONV blocks in stage-1, a downsampling block with 2 CONV blocks in stage-2, a downsampling block with 6 attention blocks in stage-3, and a downsampling block with 2 attention blocks in stage-4; see Fig.~\ref{Fig_SDTv2}. 

\vspace{-0.2cm}
\subsection{Model Compression and Pruning}

In general, pre-trained neural networks (NNs) are usually over-parameterized, leading to inefficiencies in memory footprint and inference time. 
One of the prominent solutions is pruning method, which removes redundant parameters while maintaining performance (e.g., accuracy).
Pruning has two approaches, i.e., \textit{unstructured pruning} and \textit{structured pruning}.
Unstructured pruning removes individual weights based on some importance criterion, regardless of their location, while structured pruning removes weights in regular way based on some importance criterion and their location~\cite{Ref_Cheng_SurveyPruning_TPAMI24}.
Therefore, structured pruning removes groups of weights such as channels or filters, making the compressed model more amenable to hardware acceleration~\cite{li2017pruning}\cite{Ref_Cheng_SurveyPruning_TPAMI24}. 
While pruning can significantly reduce model size and latency, most methods rely on post-pruning fine-tuning~\cite{han2015deep} to recover accuracy lost during the sparsification process. 
This reliance on retraining creates barriers for practical use in scenarios where training data is unavailable or computational resources are limited.
Toward this, recent work in ANNs explores fine-tuning-free pruning strategy~\cite{bai2023unified}, enabling efficient model compression without access to the training data. 
However, this strategy remains underexplored in the context of SNNs and SViTs, where spike-based components and dynamics impose additional constraints on pruning granularity and sparsity.

\section{The PSViT Methodology}

The objective of our work is to design a single-shot structured channel-wise pruning methodology tailored for SViT models, called \textbf{PSViT}. 
Fig.~\ref{Fig_PSViT} presents a high-level overview of our PSViT pipeline. 
The methodology takes as input a pre-trained SViT model and produces a compressed version of the same model that is structurally pruned, requires no fine-tuning, and retains strong accuracy. 
Details of PSViT pipeline are discussed in the following subsections.

\subsection{Layer-wise Sensitivity and Memory Analysis}
\label{Sec_PSViT_Sensitivity}

This step targets to understand the impact of applying structural pruning in individual network layer on the accuracy and model size.
Our layer-wise sensitivity analysis is presented as pseudo-code in Alg.~\ref{alg:sensitivity}, and described below.
\begin{itemize}[leftmargin=*]
    \item We identify candidate layers for pruning through a memory profiling step. 
    Each layer’s parameter footprint is calculated to assess its contribution to the total model size. Then, the top-$K$ layers (ranked by memory usage) are selected for pruning.
    \item To evaluate sensitivity, we perform structured pruning on each selected layer independently across a range of pruning ratios (i.e., from 1\% to 64\%). At each pruning setting, the accuracy of the pruned model is recorded and analyzed. 
\end{itemize}

\begin{figure}[t]
\centering
\includegraphics[width=0.95\linewidth]{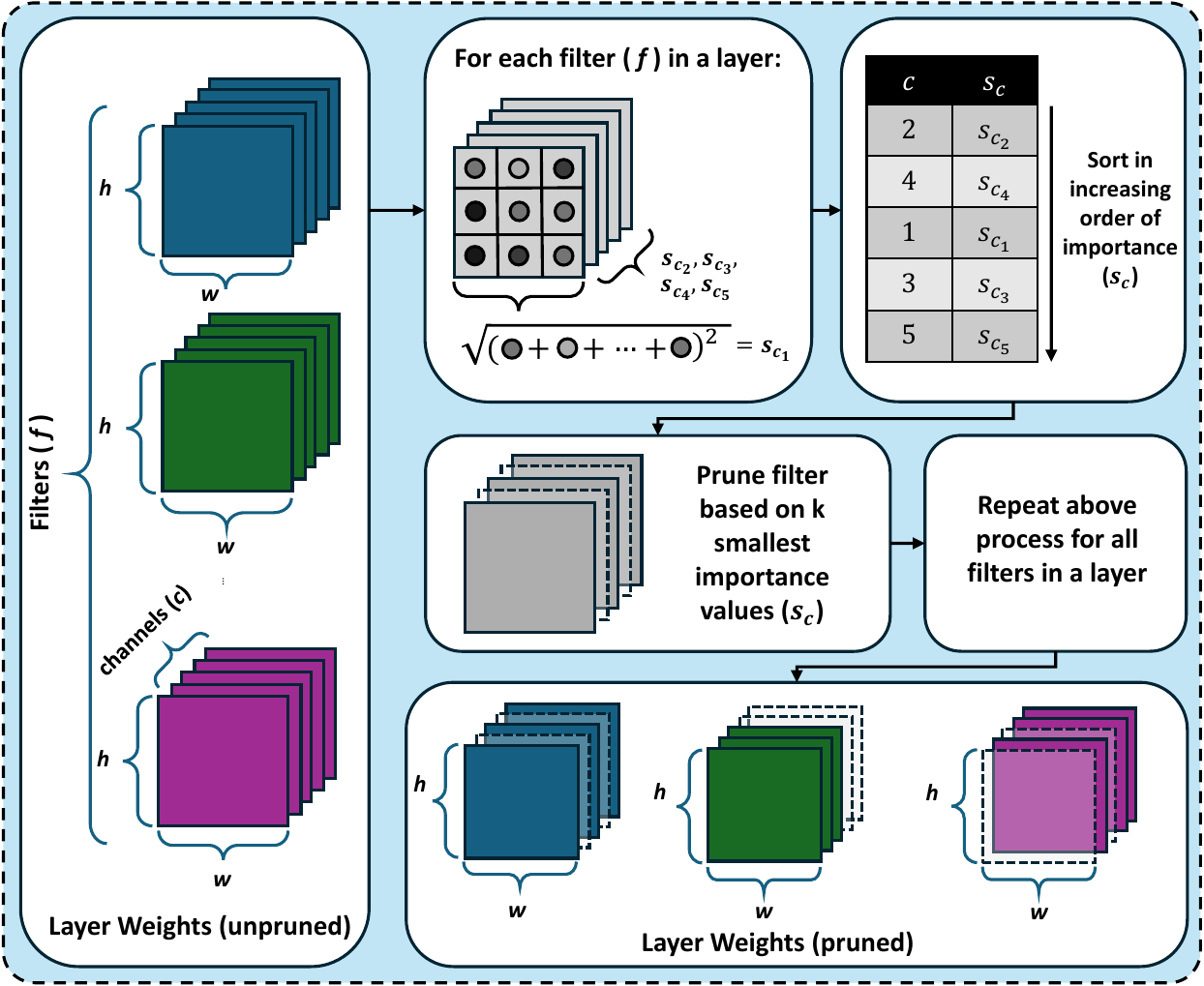}
\vspace{-0.2cm}
\caption{Our channel-wise structured pruning procedure.}
\label{fig:InChannelStructuredPruningVisual}
\end{figure}

Here, we consider channel-wise pruning, since removing channels of weights can reduce the model size significantly without changing the network topology significantly. 
Our channel-wise pruning procedure is shown in Fig.~\ref{fig:InChannelStructuredPruningVisual}, presented as pseudo-code in Alg.~\ref{alg:inchannel}, and described below. 

\begin{itemize}[leftmargin=*]
    \item For each filter \( f \) within a layer, we compute an importance score \( s_c \) for each input channel \( c \) to guide pruning. 
    \item In CONV and pointwise CONV layers, the score is defined as the square root of the sum of squared weights (see line~6 of Alg.~\ref{alg:inchannel}). For linear layers, the score simplifies to the absolute value of the scalar weight (see line~8 of Alg.~\ref{alg:inchannel}).
    These scores reflect the magnitude of contribution made by each input channel to the output of its corresponding filter.
    \item Within each filter, the input channels with the lowest \( s_c \) values, i.e., the least influential channels, are pruned according to the target pruning ratio \( r \) (see lines 5–9 and 14–18 in Alg.~\ref{alg:inchannel}). Here, we do not aggregate across filters, allowing channel importance to be assessed independently within each filter.
    \item This procedure is repeated independently for every filter within a layer, ensuring fine-grained pruning granularity. 
\end{itemize}
This structured filter-local pruning enables aggressive sparsity while preserving the hierarchical network composition. 

\begin{algorithm}[t]
\small
\caption{Layer-wise Sensitivity Analysis leveraging Channel-wise Structured Pruning}
\label{alg:sensitivity}
\begin{algorithmic}[1]
\renewcommand{\algorithmicrequire}{\textbf{INPUT:}}
\renewcommand{\algorithmicensure}{\textbf{OUTPUT:}}
\REQUIRE (1) Pre-trained SViT model (\texttt{Net}) with weights $W$; \\
\hspace*{2.3em}(2) Prunable layer list $\mathcal{L} = \{l_1, l_2, \ldots, l_N\}$; \\
\hspace*{2.3em}(3) Pruning ratios $\mathcal{R} = \{1, 2, 4, 8, 16, 32, 64\}$; \\
\ENSURE (1) Accuracy scores table $\texttt{Acc}[l][r]$; \\
\textbf{BEGIN}
\STATE Initialize accuracy table: \texttt{Acc} $\leftarrow$ empty
\FOR{each layer $l \in \mathcal{L}$}
    \STATE Reload original weights into \texttt{Net}
    \STATE Extract weight tensor $W$ and type $T$ of layer $l$
    \STATE Get number of input channels $C$
    \IF{$T$ \textbf{not in} \{\texttt{conv}, \texttt{pwconv}, \texttt{linear}\} \textbf{or} $C == 1$}
        \STATE \textbf{continue} \COMMENT{skip unsupported or depthwise layers}
    \ENDIF
    \FOR{each pruning ratio $r \in \mathcal{R}$}
        \STATE $W' \leftarrow$ \texttt{ChannelWisePrune}($W$, $T$, $r$); // from Alg. 2
        \STATE Replace weights of $l$ in \texttt{Net} with $W'$
        \STATE \texttt{Acc}[$l$][$r$] = evaluate(\texttt{Net})
    \ENDFOR
\ENDFOR
\STATE \textbf{return} \texttt{Acc}; \\
\textbf{END}
\end{algorithmic}
\end{algorithm}
\setlength{\textfloatsep}{6pt}

\begin{algorithm}[t]
\small
\caption{Our Channel-wise Structured Pruning Procedure (\texttt{ChannelWisePrune})}
\label{alg:inchannel}
\begin{algorithmic}[1]
\renewcommand{\algorithmicrequire}{\textbf{INPUT:}}
\renewcommand{\algorithmicensure}{\textbf{OUTPUT:}}
\REQUIRE (1) Layer weights $W$ of shape $F \times C \times *$; \\
\hspace*{2.3em}(2) Layer type (\texttt{conv}, \texttt{pwconv}, or \texttt{linear}); \\
\hspace*{2.3em}(3) Pruning ratio $r$; \\
\ENSURE (1) Pruned weights $W'$
\STATE Copy $W' \leftarrow W$
\STATE Get number of filters $F$ and input channels $C$
\FOR{$f = 1$ to $F$}
    \FOR{$c = 1$ to $C$}
        \IF{layer type is \texttt{conv} or \texttt{pwconv}}
            \STATE $s_c = \sqrt{\sum_{h,w} (W[f,c,h,w])^2}$
        \ELSE
            \STATE $s_c = \left| W[f, c] \right|$
        \ENDIF
    \ENDFOR
    \STATE $k = \max(1, \lfloor \frac{r}{100} \cdot C \rceil)$
    \STATE $\mathcal{I}_f =$ indices of $k$ channels with smallest $s_c$
    \FOR{each $c \in \mathcal{I}_f$}
        \IF{layer type is \texttt{conv} or \texttt{pwconv}}
            \STATE Set $W'[f,c,:,:] = 0$
        \ELSE
            \STATE Set $W'[f,c] = 0$
        \ENDIF
    \ENDFOR
\ENDFOR
\STATE \textbf{return} $W'$; \\
\textbf{END}
\end{algorithmic}
\end{algorithm}
\setlength{\textfloatsep}{6pt}

\subsection{Uniform Channel-wise Structured Pruning}
\label{Sec_PSViT_UniformPrune}

Following the layer-wise sensitivity analysis, we develop a uniform pruning configuration across layers of the network (Alg.~\ref{alg:uniform}). 
To do this, we first identify the top-$K$ most memory-consuming layers, and then apply the same pruning ratio $r$ to all of them. 
Fig.~\ref{alg:uniform} outlines this step, specifically where pruning is applied across selected layers using \texttt{ChannelWisePrune}, and the accuracy is evaluated for each pruning ratio in $\mathcal{R}$.

\begin{algorithm}[t]
\small
\caption{Our Uniform Channel-wise Structured Pruning}
\label{alg:uniform}
\begin{algorithmic}[1]
\renewcommand{\algorithmicrequire}{\textbf{INPUT:}}
\renewcommand{\algorithmicensure}{\textbf{OUTPUT:}}
\REQUIRE (1) Pre-trained SViT model (\texttt{Net}) with weights $W$;\\
\hspace*{2.3em}(2) Top-$K$ largest layers by size: $\mathcal{L} = \{l_1, \ldots, l_K\}$;\\
\hspace*{2.3em}(3) Pruning ratios $\mathcal{R} = \{1, 2, 4, 8, 16, 32, 64\}$\\
\ENSURE (1) Accuracy scores per pruning ratio: \texttt{Acc}[$r$]\\
\textbf{BEGIN}
\STATE Initialize table \texttt{Acc} $\leftarrow$ empty
\FOR{each pruning ratio $r \in \mathcal{R}$}
    \STATE Reload original weights into \texttt{Net}
    \FOR{each layer $l \in \mathcal{L}$}
        \STATE Extract weight tensor $W$ and type $T$ of layer $l$
        \STATE Get number of input channels $C$
        \IF{$T$ \textbf{not in} \{\texttt{conv}, \texttt{pwconv}, \texttt{linear}\} \textbf{or} $C == 1$}
            \STATE \textbf{continue}
        \ENDIF
        \STATE $W' \leftarrow$ \texttt{ChannelWisePrune}($W$, $T$, $r$)
        \STATE Replace weights of $l$ in \texttt{Net} with $W'$
    \ENDFOR
    \STATE \texttt{Acc}[$r$] = evaluate(\texttt{Net})
\ENDFOR
\STATE \textbf{return} \texttt{Acc}; \\
\textbf{END}
\end{algorithmic}
\end{algorithm}
\setlength{\textfloatsep}{6pt}

\vspace{-0.3cm}
\subsection{Fine-Grained Channel-wise Pruning}
\label{Sec_PSViT_FinePrune}

This step consists of performing adaptive, block-wise pruning. Alg.~\ref{alg:finegrained} captures this process. 
Starting from a uniform channel-wise pruning with ratio $r_{\text{base}}$ (line 3 of Alg.~\ref{alg:finegrained}), we test the sensitivity of each block by increasing its pruning ratio to $r_{\text{test}}$, while keeping others at $r_{\text{base}}$ (lines 4-5 of Alg.~\ref{alg:finegrained}). 
If the model retains acceptable accuracy, the elevated ratio is kept.
After the block-wise pruning evaluation, we introduce a manual override mechanism to make targeted adjustments at the individual level (lines 9-12 of Alg.~\ref{alg:finegrained}). 
This was done to better align with memory-saving goals or to recover any accuracy lost during earlier pruning stages. 
The overrides are specified through a dictionary $\mathcal{M}$ that maps layer names to custom pruning ratios (e.g., 20\%, 50\%). 
This step enables more fine-grained control than uniform or block-level pruning, and was guided by both the layer-level sensitivity trends and empirical results. 
After this step, the channel-wise filter pruning is applied with the updated pruning ratios for each layer, and the fine-grained pruned model is obtained.

\begin{algorithm}[t]
\small
\caption{Fine-Grained Channel-wise Structured Pruning}
\label{alg:finegrained}
\begin{algorithmic}[1]
\renewcommand{\algorithmicrequire}{\textbf{INPUT:}}
\renewcommand{\algorithmicensure}{\textbf{OUTPUT:}}
\REQUIRE (1) Pre-trained SViT model (\texttt{Net}) with weights $W$;\\
\hspace*{2.3em}(2) Prunable block list $\mathcal{B} = \{b_1, b_2, \ldots, b_M\}$;\\
\hspace*{2.3em}(3) Mapping blocks to layers $\mathcal{L}_b = \{b_1: [l_1, l_2], \ldots\}$;\\
\hspace*{2.3em}(4) Baseline pruning ratio $r_{\text{base}}$;\\
\hspace*{2.3em}(5) Elevated test ratio $r_{\text{test}}$;\\
\hspace*{2.3em}(6) Manual layer overrides: dictionary $\mathcal{M}[l]$\\
\ENSURE (1) Final pruned model \texttt{Net}$^\prime$\\
\textbf{BEGIN}
\STATE Initialize accuracy table: \texttt{Acc} $\leftarrow$ empty
\FOR{each block $b \in \mathcal{B}$}
    \STATE Set pruning ratio $r_l = r_{\text{base}}$ for all layers $l$ in all blocks
    \STATE Temporarily set $r_l = r_{\text{test}}$ for all $l \in \mathcal{L}_b[b]$
    \STATE Apply \texttt{ChannelWisePrune} using current config
    \STATE Evaluate model and store result: \texttt{Acc}[$b$] $\leftarrow$ accuracy
\ENDFOR
\STATE Decide final pruning ratios $r_l$ for each block $b$ based on \texttt{Acc}
\FOR{each $(l, r)$ in manual overrides $\mathcal{M}$}
    \STATE Update $r_l = r$
\ENDFOR
\STATE Apply \texttt{ChannelWisePrune} using final ratios $r_l$ for each layer
\STATE \textbf{return} pruned model \texttt{Net}$^\prime$; \\
\textbf{END}
\end{algorithmic}
\end{algorithm}
\setlength{\textfloatsep}{10pt}

\section{Evaluation Methodology}

\subsection{Experimental Setup}

To evaluate our PSViT methodology, we use a PyTorch-based implementation that leverages the SpikingJelly library~\cite{Ref_Fang_SpikingJelly_SciAdv23} for SNNs
Specifically, we use Pytorch v2.4.0 with CUDA v12.1 and Ubuntu 22.04.5 LTS OS. 
Inference and performance evaluations are executed on a single NVIDIA RTX 4090 GPU device, ensuring consistent and deterministic hardware conditions across all trials of both the original (non-pruned) and pruned models.

\textbf{Models and Datasets:}
Our baseline model is the state-of-the-art Spiking Vision Transformer v2 (\textit{SDTv2})~\cite{Ref_Yao_SpikeDrivenTransformer2_ICLR24}, which serves as the reference point for all pruning and comparison experiments. 
Specifically, we use the publicly available pre-trained SDTv2 model from the original authors. 
We reproduce training, validation, and testing using their open-source codes for SDTv2 and its default hyperparameters~\cite{Ref_Yao_SpikeDrivenTransformer2_ICLR24} on the large-scale ImageNet-1K dataset~\cite{Ref_Deng_ImageNet_CVPR09}. 
Under these settings, the baseline SDTv2 achieves 73.3\% top-1 accuracy.   

\textbf{Evaluation Metrics:}
Each model is evaluated using the ImageNet-1K dataset with a standard input shape of $(1, 3, 224, 224)$ for batch size, channels, height, and width, respectively. 
All pruning is applied post-training to a fully converged SDTv2 model, without any fine-tuning or retraining. 
This enables us to measure the effectiveness of pruning in a fine-tuning-free setting.
We evaluate both the original (non-pruned) and pruned models under identical data pre-processing and evaluation pipelines to ensure fairness. 
Here, evaluation metrics include top-1 accuracy on the ImageNet validation set, number of parameters (memory footprint), multiply-accumulate operations (MACs), and floating-point operations (FLOPs).

\section{Results and Discussion}

\begin{table*}[h]
\caption{Comparison of recent pruning methods on ImageNet. *) Results for SDTv2 and PSViT are obtained from our experiments.}
\label{Tab_Comparison}
\vspace{-0.3cm}
\footnotesize
\centering
\renewcommand{\arraystretch}{1.2}
\setlength{\tabcolsep}{6pt}
\begin{tabular}{l l c c c c c}
\hline
\textbf{Method} & \textbf{Architecture} & \textbf{Spike-Driven} & \textbf{No. Parameters [M]} & \textbf{Memory [MB]} & \textbf{Timestep} & \textbf{Accuracy [\%]} \\
\hline
\hline
\textbf{ANN} & ViT~\cite{Ref_Dosovitskiy_Transformers_ICLR21} & No & 86 & 328 & 1 & 79.7 \\
\hline
\textbf{ANN-to-SNN} 
& ResNet-34~\cite{Ref_Rathi_DeepSNN_ICLR20} & Yes & 22 & 83 & 250 & 61.5 \\
& VGG-16~\cite{Ref_Hu_FastSNN_TPAMI23} & Yes & 138 & 528 & 7 & 73.0 \\
\hline
\textbf{SViT} 
& Spikformer~\cite{Ref_Zhou_Spikformer_ICLR23} & Yes & 66.3 & 253 & 4 & 74.8 \\
& SDT~\cite{Ref_Yao_SpikeDrivenTransformer_NeurIPS23} & Yes & 66.3 & 253 & 4 & 76.3 \\
& SDTv2~\cite{Ref_Yao_SpikeDrivenTransformer2_ICLR24} & Yes & 55.4 & 211 & 4 & 73.3* \\
\hline
\textbf{Data/Token Pruning SViT} 
& STSFP~\cite{Ref_Zhou_STSFP_TCDS25} & Yes & n/a & n/a & 4 & 81.6 \\
& STATA~\cite{Ref_Zhuge_TokenSparse_ICML24} & Yes & n/a & n/a & 4 & 74.0 \\
\hline
\textbf{Structured Weight Pruning} 
& \textbf{PSViT (Ours)} & Yes & \textbf{55.4 $\rightarrow$ 43.0} & \textbf{211 $\rightarrow$ 164} & \textbf{4} & \textbf{70.3}* \\
\bottomrule
\end{tabular}
\vspace{-0.2cm}
\end{table*}
\setlength{\textfloatsep}{1pt}

\subsection{Results for Layer-wise Sensitivity Analysis}

\begin{figure}[h]
\vspace{-0.2cm}
\centering
\includegraphics[width=\linewidth]{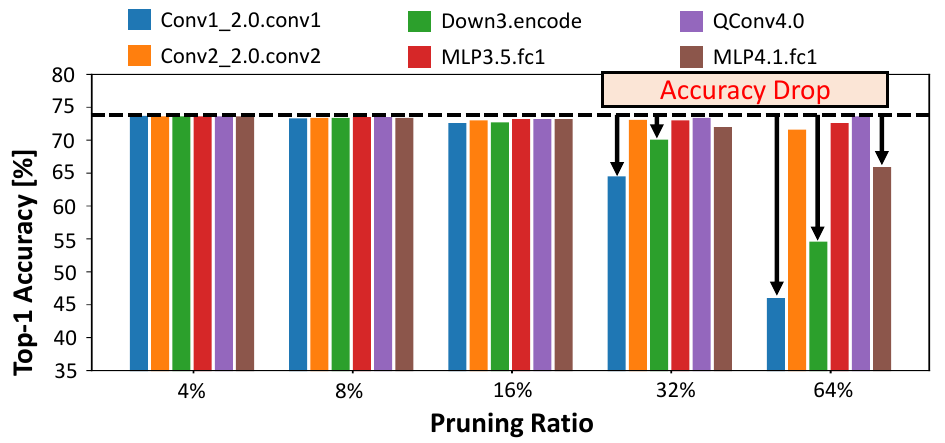}
\vspace{-0.7cm}
\caption{Results for layer-wise sensitivity analysis.}
\label{fig:layer_sensitivity_results}
\vspace{-0.5cm}
\end{figure}

During the layer-wise sensitivity analysis step, we select the top-90 largest layers from the given network (i.e., SDTv2). 
Then, we conduct the structured channel pruning on these elected layers.
Specifically, we apply independent pruning to each layer considering the pre-defined ratios (i.e., 1\%, 2\%, 4\%, 8\%, 16\%, 32\%, 64\%, 84\%, and 92\%).
Experimental results are provided in Fig.~\ref{fig:layer_sensitivity_results}, from which we make the following key observarions.
\begin{itemize}[leftmargin=*]
    \item Most network layers have $\leq$0.5\% accuracy drop, when applying channel-wise pruning up to $32\%$ pruning ratio. 
    Exceptions are observed in a few downsampling layers with $\leq$1\% accuracy degradation. 
    These results indicate that such pruning settings can be leveraged for further exploring pruning settings. 
    \item Significant accuracy drops observed beyond 64\%, particularly in early downsampling layers. It indicates that such layers are relatively sensitive to channel-wise pruning, and hence they may be exploited for aggressive pruning purpose.
    \item Downsampling layers were highly sensitive due to their role in preserving high-level spatial information. 
    On the other hand, structured pruning removes entire input channels, disrupting feature preservation in downsampling, leading to a sharp accuracy decline.
    Therefore, these layers should not be aggressively pruned, but rather pruned with small pruning ratios. 
\end{itemize} 
These observation results motivate non-uniform and layer-specific pruning configurations.

\subsection{Uniform Pruning Across Top Layers}

In this stage, structured channel-wise pruning was applied uniformly to the top-40 convolutional layers, which collectively account for approximately 75\% of the model’s total memory footprint. A fixed pruning ratio of 16\% was used for all selected layers, yielding a modest Top-1 accuracy reduction from 73.3\% to 72.3\% and achieving approximately 11.9\% memory savings. While effective in reducing model size, the results indicate that uniform pruning offers limited potential for further compression without incurring significant accuracy degradation, motivating the adoption of non-uniform, layer-specific pruning strategies in subsequent stages.

\subsection{Block-Wise Sensitivity Analysis}

\begin{figure}[h]
\vspace{-0.2cm}
\centering
\includegraphics[width=\linewidth]{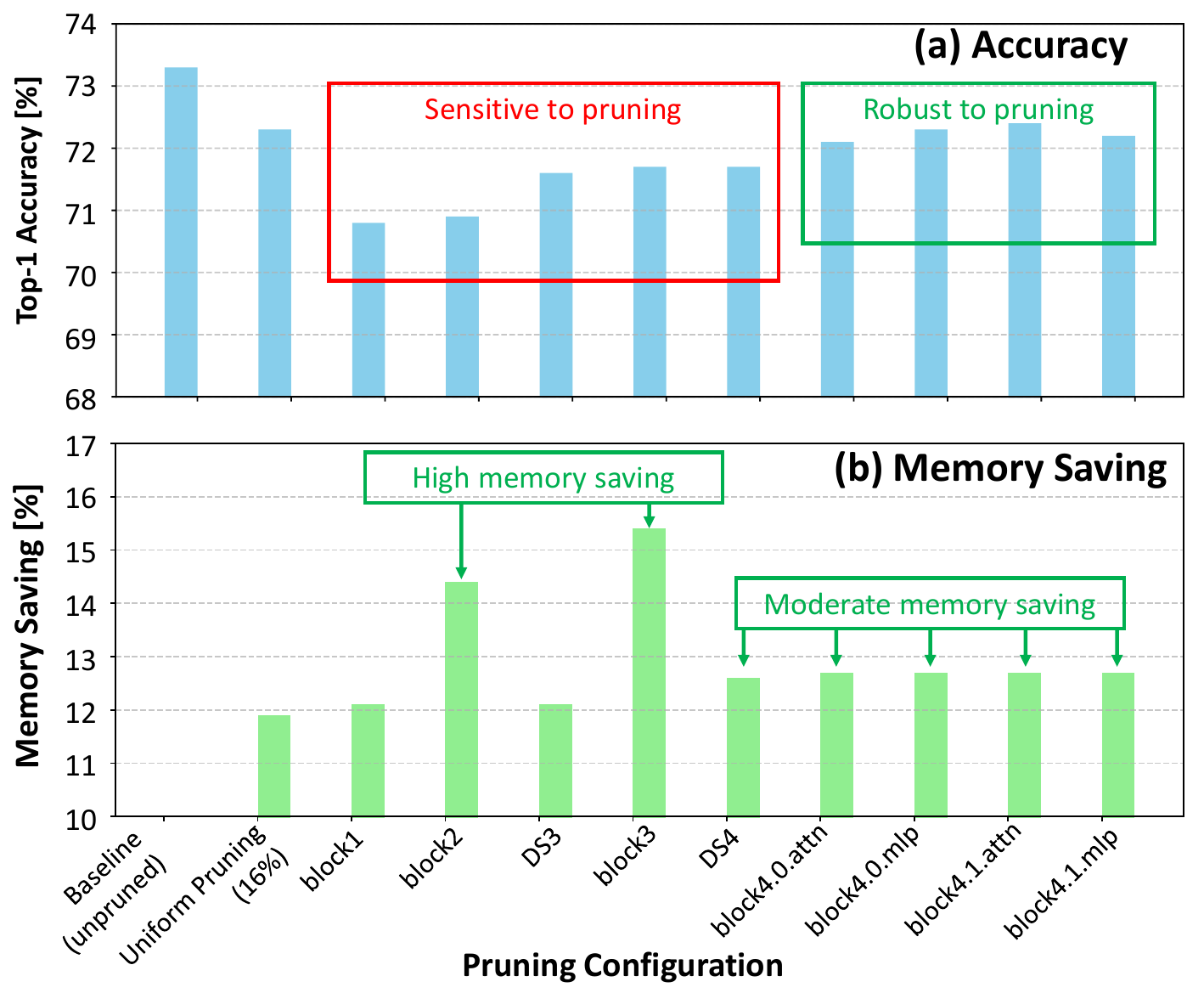}
\vspace{-0.7cm}
\caption{Results for block-wise sensitivity analysis.}
\label{fig:blockwise_sensitivity_results}
\vspace{-0.4cm}
\end{figure}

Subsequently, we performed block-wise structured pruning to refine the uniform pruning configuration. In this stage, each block (or layer group) was individually pruned at an elevated ratio of 32\%, while the remaining top-40 layers were kept at the baseline 16\% pruning level. This block-wise sensitivity analysis enabled the isolation of accuracy impacts attributable to pruning specific regions of the network. The experimental results in Fig.~\ref{fig:blockwise_sensitivity_results} reveal the following:

\begin{itemize}[leftmargin=*]
    \item Increasing pruning uniformly beyond 16\% yields diminishing returns, with accuracy losses outweighing compression benefits.
    \item Certain blocks, such as convolutional blocks 1–3 and downsampling stages 3–4, are particularly sensitive to aggressive pruning.
    \item Other components (e.g., attention and MLP) exhibit higher resilience, retaining competitive accuracy under increased pruning.
    \item Memory savings vary substantially across blocks, indicating opportunities for targeted pruning of memory-intensive blocks.

\end{itemize}

This analysis distinguishes pruning-resilient blocks that can tolerate higher sparsity without significant accuracy degradation from more sensitive regions that require conservative pruning, informing the fine-grained configuration adopted in the next stage.

\subsection{Best Trade-Off Configurations}

\begin{figure}[t]
\vspace{-0.2cm}
\centering
\includegraphics[width=\linewidth]{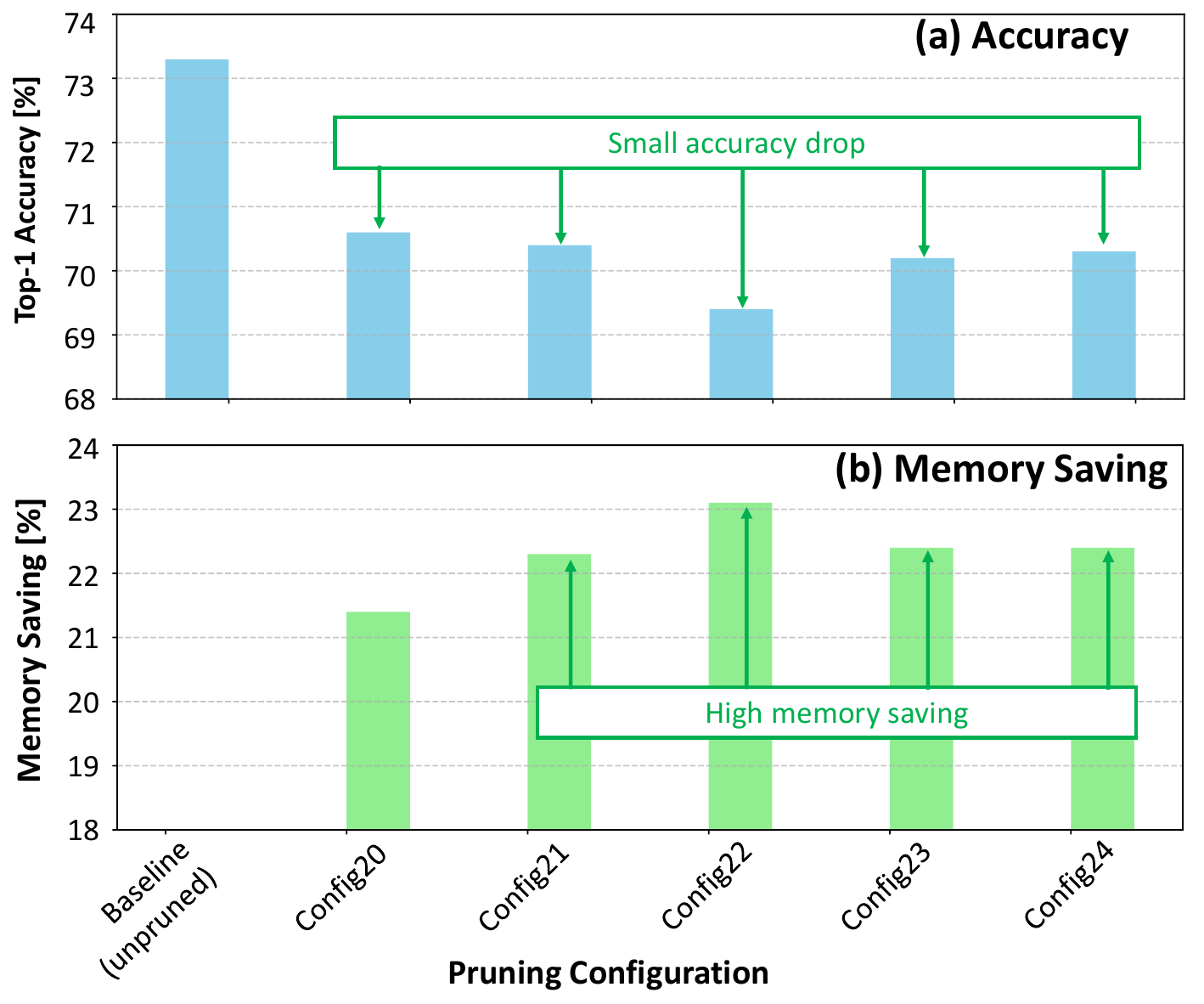}
\vspace{-0.7cm}
\caption{Results of pruning configurations with the best trade-off between accuracy and memory saving.}
\label{fig:finegrained_sensitivity_results}
\vspace{0.3cm}
\end{figure}

Based on the previous steps, we can identify the final single-shot pruning candidates, which are presented in Table~\ref{Tab_FineGrainedConfigs}-\ref{Tab_ConfigPerformance} and Fig.~\ref{fig:finegrained_sensitivity_results}.
From these results, we conclude that the potential final pruning configurations include: (1) \textbf{Config 21} with 70.4\% top-1 accuracy and 22.3\% memory saving, and (2) \textbf{Config 24} with 70.3\% top-1 accuracy and 22.4\% memory saving. These results are obtained without any post-pruning fine-tuning or reconstruction.
These significant reductions in memory savings while having high accuracy preservation come from our systematic pruning approach that applies aggressive pruning for non-sensitive layers and less aggressive pruning for sensitive layers.
Furthermore, we evaluate the impact of fine-tuning on the \textbf{Config 24} model. 
The experimental results indicate that the accuracy of the \textbf{Config 24} model recovers to 72.8\%, approaching close to baseline accuracy (73.3\%).
These results highlight the potential of the PSViT methodology to achieve final accuracy very close to the baseline due to its effective pruning strategy.

\begin{table}[t]
\caption{Fine-grained pruning configurations showing the specific layers modified from the baseline setup.}
\label{Tab_FineGrainedConfigs}
\small
\centering
\renewcommand{\arraystretch}{1.2}
\setlength{\tabcolsep}{6pt}
\resizebox{\linewidth}{!}{
\begin{tabular}{c l}
\hline
\textbf{Config} & \textbf{Modified Layers and Pruning Ratios} \\
\hline
20 & ConvBlock2\_1.0.conv2: 40\%, ConvBlock2\_2.0.conv2: 40\%, \\
   & downsample4.encode\_conv: 20\% \\
21 & Config 20 + ConvBlock2\_1.0.conv2: 50\%, ConvBlock2\_2.0.conv2: 50\% \\
22 & Config 21 + ConvBlock2\_1.0.conv2: 60\%, ConvBlock2\_2.0.conv2: 60\% \\
24 & Config 22 + downsample3.encode\_conv: 20\%, ConvBlock1\_2.0.conv2: 20\% \\
\hline
\end{tabular}}
\end{table}
\setlength{\textfloatsep}{1pt}
\vspace{-0.4cm}

\vspace{0.3cm}
\subsection{Ablation Study and Further Discussion}

\subsubsection{\textbf{Impact of Compute Cost}}
Table~\ref{Tab_ConfigPerformance} compares multiple pruning configurations in terms of Top-1 accuracy, memory savings, and computational cost. The baseline SDTv2 model achieves 73.3\% accuracy with no pruning, a memory footprint of 20.98\,G MACs, and 42.07\,G FLOPs. All pruned configurations yield substantial reductions in memory and computational requirements, with memory savings ranging from 21.4\% to 23.1\% and corresponding decreases in MACs and FLOPs. Among these, configuration~24 achieves a favorable trade-off, preserving 70.3\% accuracy while reducing MACs and FLOPs by approximately 17\% compared to the baseline. In contrast, configuration~22 offers the highest memory savings (23.1\%) but with a larger accuracy drop to 69.4\%. These results show the importance of selecting pruning configurations that balance compression with minimal accuracy degradation.

\begin{table}[t]
\caption{Performance of pruning configurations in terms of top-1 accuracy, memory savings, computational cost, and estimated energy consumption.}
\label{Tab_ConfigPerformance}
\footnotesize
\centering
\begin{tabular}{c c c c c c} 
\hline
\textbf{Config} & \begin{tabular}[c]{@{}c@{}} \textbf{Acc-1} \\ \textbf{[\%]} \end{tabular} & \begin{tabular}[c]{@{}c@{}} \textbf{Memory} \\ \textbf{Saving [\%]} \end{tabular} & \begin{tabular}[c]{@{}c@{}} \textbf{MACs} \\ \textbf{[G]} \end{tabular} & \begin{tabular}[c]{@{}c@{}} \textbf{FLOPs} \\ \textbf{[G]} \end{tabular} & \begin{tabular}[c]{@{}c@{}} \textbf{Estimated} \\ \textbf{Energy [mJ]} \end{tabular} \\
\hline
Baseline & 73.3 & 0.0 & 20.98 & 42.07 & 6.32 \\
20       & 70.6 & 21.4 & 17.75 & 35.61 & 5.79 \\
21       & 70.4 & 22.3 & 17.39 & 34.87 & 5.74 \\
22       & 69.4 & 23.1 & 17.02 & 34.14 & 5.69 \\
24       & 70.3 & 22.4 & 17.30 & 34.71 & 5.67\\
\hline
\end{tabular}
\end{table}
\setlength{\textfloatsep}{6pt}

\subsubsection{\textbf{Further Discussion}}
PSViT does not directly apply the existing pruning methods from the ANN domain (i.e., ViTs) to SViT models, as PSViT performs original studies and observations directly on SViTs, and then applies its pruning methodology tailored for the investigated models. 
Specifically, PSViT establishes a practical structured pruning design pipeline suitable for spiking neuronal dynamics, which differs significantly from ANNs (i.e., ViTs), including the timestep computation and spiking dynamics in attention, linear, and CONV layers. 
PSViT also performs sensitivity analysis to identify layers that are highly tolerant to pruning, then employs appropriate channel-wise pruning rates to compress these layers. 
Moreover, this SNN-suitable and training-free pruning strategy is not covered by the existing SViT pruning studies, since they mainly focus on input data/token sparsification, unstructured weight sparsity, or iterative retraining-based pruning, hence highlighting the novel contributions of our PSViT methodology.
With its single-shot pruning approach, PSViT provides a solution for avoiding excessive retraining overhead of iterative pruning methods. 

\section{Conclusion}

We introduce a novel PSViT methodology that enables a structured channel-wise pruning with single-shot approach for SViTs. 
By combining layer-wise sensitivity analysis, uniform channel-wise pruning of the most memory-intensive layers, and fine-grained block-layer and layer-level adjustments, our PSViT delivers substantial reductions of model size and FLOPs, while maintaining competitive ImageNet-1K accuracy, even at aggressive pruning ratios. 
Specifically, PSViT compresses SDTv2 model and achieves 22.4\% model size reduction and 17.5\% smaller FLOPS, while maintaining competitive ImageNet-1K accuracy within 3\% (70.3\% without fine-tuning and 72.8\% with fine-tuning) compared to 73.3\% from the original non-pruned model.
The results show that SViTs can be compressed effectively, improving their viability for embedded deployments.

\section*{Acknowledgment}
This work was partially supported by the NYUAD Center for Interacting Urban Networks (CITIES), funded by Tamkeen under the NYUAD Research Institute Award CG001.

\bibliographystyle{IEEEtran}
\bibliography{bibliography}

\end{document}